\documentclass[journal,twoside]{IEEEtran}
\usepackage{float}
\usepackage{graphicx}
\usepackage{subfigure}
\usepackage{cite}
\usepackage{picinpar}
\usepackage{amsmath,amssymb,amsfonts}
\usepackage{url}
\usepackage[utf8]{inputenc}
\usepackage[T1]{fontenc}
\usepackage{colortbl}
\usepackage{soul}
\usepackage{multirow}
\usepackage{pifont}
\usepackage{color}
\usepackage{alltt}
\usepackage[hidelinks]{hyperref}
\usepackage{enumerate}
\usepackage{siunitx}
\usepackage{epstopdf}
\usepackage{pbox}
\usepackage{caption}
\usepackage{array}
\usepackage{booktabs}
\usepackage{nomencl}
\usepackage{stfloats} 
\usepackage{diagbox} 
\usepackage[ruled,linesnumbered]{algorithm2e}
\usepackage{amsthm}
\usepackage{enumitem}
\usepackage{xcolor}
\usepackage{soul}

\def\BibTeX{{\rm B\kern-.05em{\sc i\kern-.025em b}\kern-.08em
    T\kern-.1667em\lower.7ex\hbox{E}\kern-.125emX}}
\IEEEpubid{\begin{minipage}{\textwidth}\ \\[34pt] \centering
		\copyright 2026 IEEE. Personal use of this material is permitted. Permission from IEEE must be obtained for all other uses... Accepted for publication in CSEE Journal of Power and Energy Systems.
\end{minipage}}
\begin{document}

\title{Electric Vehicle User Charging Behavior Analysis Integrating Psychological and Environmental Factors: A Statistical-Driven LLM based Agent Approach}
\author{Chuanlin Zhang, \IEEEmembership{Senior Member, IEEE}, Junkang Feng, Chenggang Cui, \IEEEmembership{Member, IEEE},  Pengfeng Lin, \IEEEmembership{Member, IEEE}, Hui Chen, \IEEEmembership{Member, IEEE}, Yan Xu, \IEEEmembership{Senior Member, IEEE}, Amer M. Y. M. Ghias, \IEEEmembership{Senior Member, IEEE}, Qianguang Ma, Pei Zhang, \IEEEmembership{Fellow, IEEE}
\thanks{This work is supported in part by the National Natural Science Foundation of China under Grant 62233006.}
\thanks{C. Zhang, J. Feng and C. Cui (corresponding author, e-mail: cgcui@shiep.edu.cn) are with the Intelligent Autonoumous Systems Lab, Shanghai University of Electric Power, Shanghai, China 200090.}
\thanks{P. Lin is with the School of Electrical Engineering, Shanghai Jiaotong University, Shanghai, China 200240.}
\thanks{A. M. Y. M. Ghias and Y. Xu are with the School of Electrical and Electronic Engineering, Nanyang Technological University, Singapore 639798.}
\thanks{Q. Ma is with the Psychological Consultation Center, East China University of Political Science and Law, Shanghai, China 201620.}
\thanks{P. Zhang is with the School of Electrical and Information Engineering, Tianjin University, Tianjin, China 300072.}}
\maketitle

\begin{abstract}
With the growing adoption of electric vehicles (EVs), understanding user charging behavior has become critical for grid stability and transportation planning. This study investigates the behavioral heterogeneity of EV taxi drivers by analyzing the interaction between psychological traits and situational triggers within dynamic travel contexts. Leveraging large language models (LLMs) as a core simulation tool, a novel framework with statistical enhancement is developed to replicate and analyze the charging behaviors of taxi drivers. LLMs simulate personalized decision-making processes by leveraging natural language reasoning and role-playing capabilities, accounting for factors such as time sensitivity, price awareness, and range anxiety. Simulation results indicate that the framework reliably reproduces real-world charging behaviors across multiple urban environments. his fidelity arises from integrating statistical priors into the reasoning process, allowing the model to anchor its decisions in empirical behavioral patterns. Further analysis highlights the joint influence of environmental and psychological variables on charging decisions and reveals the heterogeneity of different user groups. The findings provide new insights into EV user behavior, offering a foundation for optimizing charging infrastructure, informing energy policy, and advancing the integration of EV behavioral models into smart transportation and energy management systems.
\end{abstract}

\begin{IEEEkeywords}
EV user charging behavior, large language model, AI agent, behavioral heterogeneity, grid stability.
\end{IEEEkeywords}

\section{Introduction}
\label{sec:introduction}
\IEEEPARstart{R}{apidly} increasing adoption of EVs poses significant challenges for both modern power grids and urban transportation systems, particularly as their large-scale integration accelerates \cite{9215181}. The diverse charging behaviors of EV users, encompassing variations in frequency, timing, location, and personal preferences, introduce substantial uncertainties into grid operations\cite{wolinetz2018simulating}. These demand fluctuations are particularly pronounced during peak hours, when the unpredictability and heterogeneity of charging patterns intensify load pressures, increasing the risk of overloads and operational disruptions that threaten grid stability and economic efficiency \cite{9862537}. The inherent heterogeneity and randomness in charging behaviors make traditional load forecasting and scheduling methods inadequate, underscoring the difficulty of understanding the underlying patterns and drivers of EV user charging behavior \cite{steinbach2024grid}.

EV user charging behavior is influenced by psychological factors such as range anxiety, price sensitivity, and charging satisfaction, along with environmental variables including time constraints, station availability, and electricity price fluctuations \cite{you2024unraveling}. These interactions complicate the prediction of user behavior and introduce significant variability into charging decisions in dynamic conditions. Traditional approaches, including surveys \cite{pareschi2020travel}, field observations \cite{lee2020exploring}, and statistical analyses \cite{cui2022battery}, have provided a foundation for understanding user behavior at a macro level but are limited by scalability, high data acquisition costs, and a lack of real-time applicability \cite{li2023electric}. Advanced methods such as data mining \cite{siddique2022data}, stochastic modeling \cite{fotouhi2019general}, and agent-based modeling have been adopted to gain deeper insights into user behavior \cite{chaudhari2018agent}. Among these, agent-based modeling can simulate interactions between individual users and charging infrastructure, uncovering heterogeneous behavior patterns. However, these techniques often fail to account for the psychological complexity and environmental sensitivity that influence charging decisions \cite{yaghoubi2024systematic}.

Existing literature has shown that psychological factors and behavioral heterogeneity substantially complicate the modeling and decision-making of EV user charging behavior in dynamic contexts. In \cite{xu2020mitigate}, the critical role of range anxiety is emphasized, along with mitigation strategies designed to alleviate driver concerns. Price sensitivity is also identified as a major determinant of user behavior, particularly influencing cost-related decisions \cite{visaria2022user}. The interaction between psychological traits and situational variables—and the necessity of longitudinal data to capture evolving user dynamics—is demonstrated in  \cite{bailey2024causal}. Behavioral heterogeneity introduces further complexity by reflecting diverse user preferences and usage patterns. For example, distinct subgroups of taxi drivers have been identified based on unique charging routines, highlighting the value of behavioral segmentation \cite{cai2025heterogeneity}. Additional studies classify users by their sensitivity to cost, time, and distance to charging stations, offering deeper insight into user-specific priorities \cite{yang2024analyzing}. Moreover, a binary logistic regression model incorporating contextual variables such as service quality and parking duration has expanded the analytical understanding of decision variability \cite{wang2021electric}. Despite these advances, current frameworks often rely on static classification methods that fail to account for dynamic adjustments in user behavior arising from the interaction between psychological traits and external environmental variables.

The rapid development of LLMs offers transformative potential for addressing the complexities of modeling EV user charging behaviors \cite{mihalcea2024developments}. Outperforming traditional methods in generalization, reasoning, and contextual understanding, LLMs are well-suited to capturing the dynamic interactions between psychological traits, behavioral heterogeneity, and environmental variables. The successful application of the RecAgent simulator in recommendation systems shows that LLMs can adapt to the evolving dynamics of user behaviors and preferences \cite{wang2024survey}. Similarly, LLM-based agents have been used to simulate emotional responses and social interactions in online communities, revealing the mechanisms of information dissemination and group formation \cite{gao2023s3}. Role playing techniques further enhance LLMs’ ability to emulate diverse personas and decision-making scenarios. Shanahan et al. show that LLMs can generate responses aligned with defined role attributes and behaviors through carefully crafted dialogue prompts \cite{shanahan2023role}. While LLMs have demonstrated adaptability in recommendation systems and social interaction modeling, challenges remain in EV user behavior modeling. EV decision-making involves dynamic interactions between psychological states and environmental factors while requiring decision consistency. To fully leverage LLMs' potential, they must be optimized to capture psychological dynamics, contextual interactions, and improve decision consistency and environmental adaptability for uncovering the mechanisms driving EV user behaviors \cite{hussain2024tutorial}.

This study proposes a dynamic modeling framework based on LLMs to analyze the evolving charging behaviors of EV users across diverse scenarios. By integrating users’ psychological traits with environmental variables, the framework enhances both the interpretability and predictive performance of user behavior models. The results reveal the intricate interplay of psychological traits, behavioral heterogeneity, and situational triggers in shaping EV user decisions. This approach provides  profound insights into the heterogeneity of charging behaviors and offers  scientific support for optimizing dynamic pricing strategies, designing efficient charging station layouts, and enhancing grid demand response capabilities. Compared with existing related works, the main contribution of this paper is threefold:
\begin{itemize}
	\item A LLM based Agent model framework is proposed, encompassing multiple modules, including action, property, rule, perception, and environment, to achieve sophisticated and adaptive simulations of electric vehicle user behavior.
	
	\item Integrating statistical analysis EV behavior results into the LLM, the framework dynamically retrieves relevant statistical information during decision\-making, significantly enhancing the decision accuracy and reliability of the LLM model.

    \item The framework leverages LLMs’ contextual reasoning and role-playing capabilities to simulates user behavior, generates text-based explanations and quantitatively analyzes behavioral triggers to uncover the drivers behind user decisions. 
\end{itemize}  

The remainder of the paper is organized as follows. Section II introduces the problem description and preliminaries. Section III details the proposed LLM-based agent framework and its methodological underpinnings. Section IV presents experimental results and analysis. Section V presents the conclusion of this study and discusses avenues for future work.

\section{Problem Description and Preliminaries}

\subsection{EV Charging Behavior Model}
To simulate EV charging behavior, an agent-based model can be constructed using a five-element framework \(\{p, a, r, e ,t\}\) \cite{wallace2015assessing}, where \(p\) represents the agent’s attributes, such as age and gender, \(a\) stands for the agent's actions, like charging decisions, \(r\) refers to the rules guiding these behaviors, and \(e\) represents the environmental influences, and \(t\) is the time factor. The model aims to capture the complexity of EV charging behavior through a  
function \(f\) which maps these elements into a coherent model of EV charging behavior.:

\begin{equation}
a = f(p, r, e, t)
\end{equation}

Within this framework, the first three elements \(p\), \(a\), and \(r\) are crucial for defining the EV user behavior agent, while the last two elements \(e\) and \(t\) define the spatiotemporal context in which these behaviors occur. The
function \(f\) maps these elements into a comprehensive and dynamic representation of EV charging behavior, taking into account individual differences and variations in the environmental and temporal contexts.

\subsection{LLM based AI Agent}
An LLM based AI agent can be represented as a five part tuple 
\begin{equation}
U = (L, O, P, M, A, R)
\end{equation}
where \( U \) denotes the LLM based AI agent, \( L \) stands for the Large Language Model, \( O \) represents the Objective, \( P \) represents the Perception, \( M \) denotes Memory, \( A \) indicates Action, and \( R \) stands for Rethink \cite{cheng2024exploring}.

The integration of LLM based agents into EV Charging Behavior models has the potential to create agents capable of simulating behavior in complex conditions. These agents possess adaptability, interpretability, zero-shot learning, common-sense reasoning, and can handle various situations. Despite these advantages, the conventional LLM based agent framework may not be directly applicable for modeling electricity user behavior. Therefore, a customized framework is necessary to enable these agents to perform complex tasks.

\section{LLM-Based Agent Framework Design}

\begin{figure*}[ht]
\centering
\includegraphics[width=1.0\textwidth]{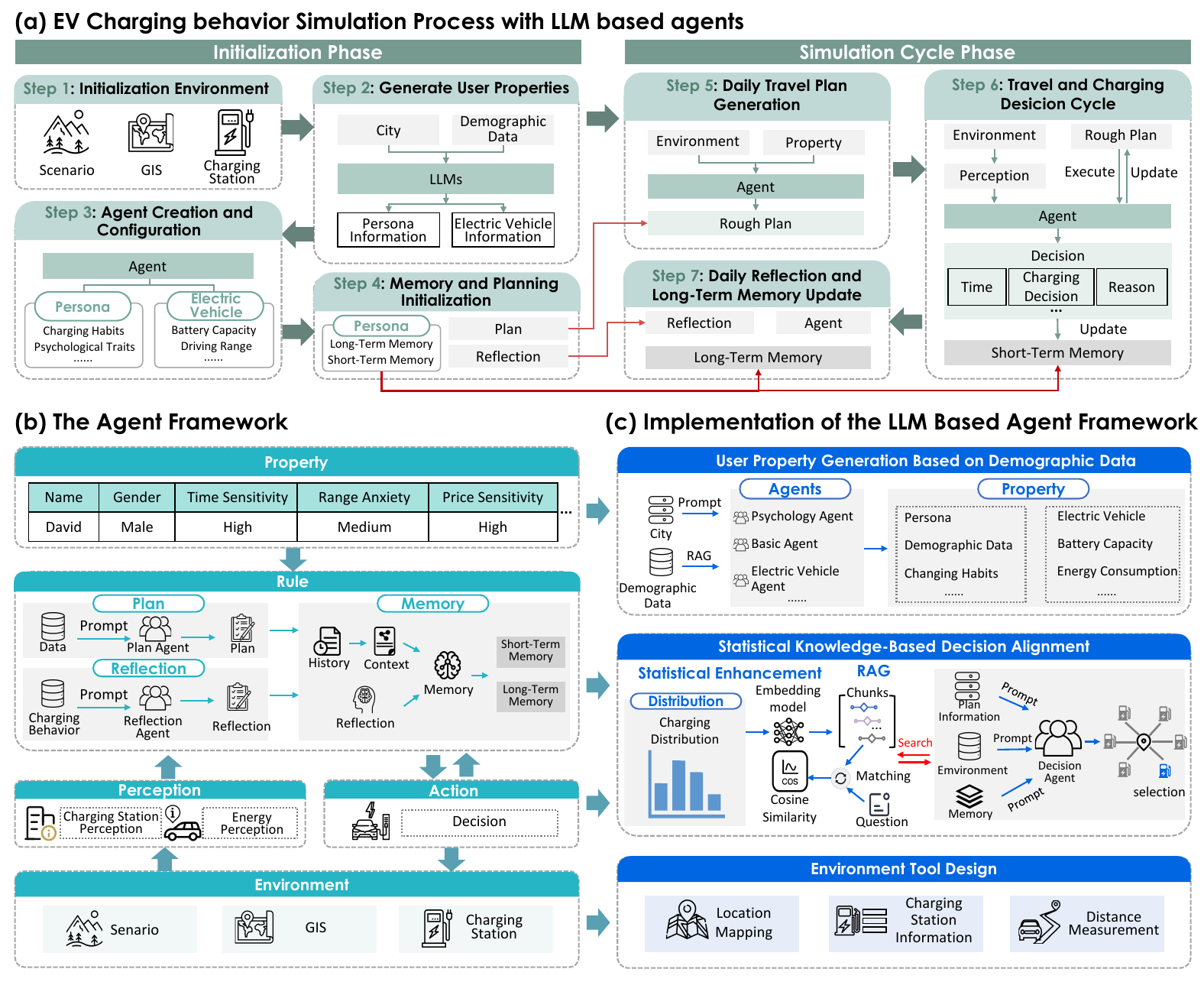}
\caption{EV charging behavior simulation process with LLM-based agents. (a) EV charging behavior simulation process with LLM-based Agents:
It illustrates the comprehensive simulation process of EV charging behavior using LLM-based agents, encompassing memory initialization, dynamic plaanning, reflection, and long-term memory updates, (b) The agent framework:
It illustrates the detailed inner structure of an agent, including its properties, planning, memory, and action flow, in response to environmental inputs, (c) Implementation of the LLM-based agent framework:
It depicts how LLMs are utilized to integrate user properties, align statistical knowledge for decision-making, and design the environmental tools.}
\label{fig:Decision-Making}
\end{figure*}

\subsection{LLM-based Agent Framework}
To enhance the accuracy of simulating EV charging behaviors, this paper presents an LLM-based agent framework, which integrates agent-based modeling and LLMs. The framework is designed to provide deeper insights and more precise predictions of EV charging behaviors. The decision-making process is formalized as follows:
\begin{equation}
a = \theta_L(p, r, P, e, t)
\end{equation}
where \(\theta_{L}\) represents the response function of the large language model. This function synthesizes all input parameters, leveraging the reasoning capabilities of LLMs to simulate user decision-making in realistic charging scenarios.The proposed framework is composed of seven modules, each playing a critical role in simulating realistic EV charging behaviors. The action module $(a)$ oversees charging decisions, station selection, and charging duration to ensure operational realism. The LLM module $(L)$ serves as the core, processing environmental changes and leveraging reasoning capabilities to enable personalized, rule-driven charging behaviors. The property module $(p)$ incorporates user attributes and EV specifications, ensuring that decisions align with user preferences and vehicle performance. The rule module $(r)$ dynamically adjusts decision logic using LLM reasoning, adapting to changing conditions. The perception module $(P)$ processes real-time data on travel, energy levels, and station availability to support informed decision-making. The environment module $(e)$ replicates real-world charging conditions, while the time module $(t)$ manages temporal factors, reflecting realistic schedules and energy needs. Together, these modules form a robust framework for analyzing user behaviors and optimizing charging systems (See Fig.~\ref{fig:Decision-Making}(b)). These modules collectively form a versatile system for a comprehensive understanding of EV user behavior.

\subsubsection{Action Module Design}
The action module systematically describes EV user charging behavior, encompassing key dimensions such as user decision-making, station selection, target state of charge (SoC), and the reason for charging. The module is expressed as: 
\begin{equation}
a = (U_d, S_s, S_e, E_x)
\end{equation} 
where \( U_d \) represents the user’s decision to initiate charging, \( S_s \) indicates the selected charging station, \( S_e \) specifies the target SoC, and \( E_x \) describes the reason for starting the charging session.

\subsubsection{Property Module Design}
The property module describes the key attributes influencing EV charging behavior and is divided into two components: user properties and EV properties.

The user properties are represented as:
\begin{equation}
p_u = (D_i, E_c, P_s, C_h)
\end{equation}
where $D_i$ denotes demographic information including foundational personal attributes such as age, gender, and occupation derived from regional statistical distributions, $E_c$ represents economic conditions such as income levels and financial status relative to the regional demographic context, $P_s$ reflects psychological traits consisting of multiple sensitivities including price and time sensitivity among others quantified into discrete levels such as high, medium, or low to serve as reasoning anchors, and $C_h$ describes charging habits encompassing established behavioral routines such as preferred charging times and location choices derived from the assigned roles of the agents.

The EV properties are represented as:
\begin{equation}
p_{EV} = (M_f, M_d, B_c, R_g, E_c)
\end{equation}
where \( M_f \) refers to the manufacturer, \( M_d \) indicates the model, \( B_c \) specifies the battery capacity, \( R_g \) represents the driving range, and \( E_c \) denotes energy consumption.

\subsubsection{Rule Module Design}
The rule module is the core component of the LLM-based agent framework, designed to define the rules governing EV charging decisions. The module is represented as the following tuple:
\begin{equation}
r = (R_p, R_m, R_r)
\end{equation}
where \( R_p \) denotes the planning submodule, \( R_m \) represents the memory submodule, and \( R_r \) corresponds to the reflection submodule. These components work together to enable the agent to plan, store, and retrieve data, as well as optimize charging decisions by reflecting on past behaviors.

\paragraph{Planning Submodule}
The planning submodule utilizes LLM to generate comprehensive daily activity, travel, and charging plans, effectively simulating the real-world  EV user behavior. This module consists of three planning components: event planning, which schedules work, shopping, and leisure activities based on the user’s daily behavior and role characteristics; travel planning, which determines travel times, plans routes, and estimates distances according to scheduled activities; and charging planning, which generates charging schedules based on SoC and travel requirements.

\paragraph{Memory Submodule} 
The memory submodule serves as the agent’s knowledge repository, managing historical charging data to support more precise decision-making processes. It includes two components: short-term memory and long-term memory. Short-term memory stores recent charging records. Long-term memory retains extended charging data, analyzing user habits and trends to inform future planning and decision-making.

\paragraph{Reflection Submodule}
The reflection submodule enables the agent to refine its decision-making process and enhance the accuracy of EV charging behavior simulations by addressing three key aspects. Plan reflection evaluates whether the executed charging decisions align with the user’s needs and expectations. User satisfaction reflection analyzes satisfaction levels regarding charging time, station selection, charging amount, power, and cost. Property reflection verifies if the decisions are consistent with the user’s behavior patterns, personal attributes, and the characteristics of the EV.

\subsubsection{Perception Module}
The perception module gathers data from travel, energy, and charging stations. It can be structured as a tuple:
\begin{equation}
P = \{P_t, P_e, P_{cs}\}
\end{equation}
where \(P_t\) represents travel perception, \(P_e\) corresponds to energy perception, and \(P_{cs}\) refers to charging station perception. Travel perception collects data on travel routes, distances, and estimated times to optimize charging decisions. Energy perception focuses on monitoring the vehicle’s SoC and energy consumption rates. Charging station perception provides detailed information on charging stations, including their geographic locations, charging prices, and power capacities.

\subsubsection{Environment Module}
The environment module consists of three key elements: charging scenarios, geographic information, and charging station data, represented as the following tuple:
\begin{equation}
e = \{e_s, e_g, e_c\}
\end{equation}
where \(e_s\) represents the charging scenario, \(e_g\) denotes the geographic information, and \(e_c\) refers to the charging station details. These data inputs are sourced from the Gaode Map API\footnote{\url{https://lbs.amap.com/api}}, which provides real-time traffic congestion and station status information for simulation purposes.

\subsection{Implementation of the LLM-based Agent Framework}
The implementation of the LLM  based agent framework for simulating EV charging behavior involves a comprehensive integration of various modules to create a robust and dynamic system. Each module plays a critical role in ensuring the agent can accurately predict and optimize EV charging processes by modeling user preferences, personalities, and behaviors (See Fig.~\ref{fig:Decision-Making}(c)). In what follows, a detailed description of the implementation for these modules is presented.

\subsubsection{User Property Generation Based on Demographic Data}
The generation of user and EV properties is accomplished through a multi-agent framework that leverages Retrieval-Augmented Generation (RAG) to integrate city data, demographic distributions, and domain-specific knowledge. Specifically, the Base Agent generates foundational demographic information ($D_i$) and economic conditions ($E_c$) by querying contextual inputs such as user location and regional statistics. Subsequently, based on the generated personal background, the LLM utilizes specific prompts to generate charging habits ($C_h$) and psychological traits ($P_s$), such as price sensitivity, range anxiety, and time sensitivity. This approach ensures that the agents' behavioral and psychological properties are logically consistent with their socioeconomic backgrounds and grounded in empirical foundational data.

Through this approach, multiple specialized agents are employed to generate comprehensive user properties \(p_u\) and EV properties \(p_{EV}\). These generated properties are then used as input conditions for downstream decision-making simulations and behavioral modeling tasks.

\subsubsection{Charging Decision Generation based on Probability Distribution Enhanced Retrieval}
To replicate EV user charging behavior, this study introduces a method for generating charging decisions using probability distribution enhanced retrieval. Charging behavior feature distributions are translated into natural language descriptions, and enhanced retrieval accesses multiple distributions. These descriptions, combined with user attributes and environmental contexts, are processed by a LLM to produce charging decisions aligned with statistical distributions.
\paragraph{Feature Statistical Data Segmentation}
The charging behavior feature statistical dataset $\Delta$ is divided into multiple clearly defined segments, represented as:
$\Delta = \{\delta_1, \delta_2, \ldots, \delta_n\}$
where each segment $\delta_i$ represents specific statistical data of a charging behavior feature. 
\paragraph{ Probability Distribution of Feature Segments}
The probability distribution of each segment feature is described by a probability density function $\rho(a)$, which characterizes the distribution of the feature over different intervals. For example, for the distribution of the initial charging SoC, each segment’s statistical interval $\text{SoC}_i$ can be represented as:
\begin{equation}
\text{SoC}_i = [\alpha_i, \beta_i], \quad i = 1, 2, \dots, n
\end{equation}
and the probability density function is defined as:
\begin{equation}
\rho(\delta) = \{\rho_i \mid \delta \in [\alpha_i, \beta_i],\ i = 1, 2, \dots, n\}
\end{equation}
where $\rho_i$ represents the probability weight for the feature value in the interval $[\alpha_i, \beta_i]$.

\paragraph{Natural Language Representation of Feature Statistics}
The probability distribution of features is divided into natural language description segments according to intervals. Each segment includes the feature name, interval range, and corresponding probability. For example, if the property ``charging initial SoC” has a probability of $\rho=0.5$ within the interval $[0.4, 0.5)$, the corresponding natural language description is:

``\textit{The probability of charging initial SoC is between 40\% and 50\% is 50\%.}”

\paragraph{Statistical Feature Query}
For a statistical feature query $\varphi \in \Phi$, the corresponding data subset is defined as \cite{zhao2024retrieval}:
\begin{equation}
\text{Dep}(\phi) = \{\delta \mid \delta \in \Delta,\ I_\phi(\delta) = 1\}
\end{equation}
where the indicator function $I_\varphi(\delta)$ is defined as:
\begin{equation}
I_\phi(\delta) =
\begin{cases}
1, & \text{if } \delta \text{ is required to answer } \phi \\
0, & \text{otherwise}
\end{cases}
\end{equation}

Based on a feature query model $\psi: \Phi \rightarrow \mathcal{P}(\Delta)$, the natural language description segments corresponding to the statistical query are represented as:
$\chi = \psi(\varphi) \cap \Delta$
where $\chi$ represents the feature description segments required to answer the query $\varphi$.

\paragraph{Charging Behavior Decision Generation}
The decision generation process is extended to include feature statistical data relevant to EV user charging behavior, formalized as:
\begin{equation}
	a = \theta_L(p, r, e, t, P, \chi)
\end{equation}
where this extension explicitly incorporates modules for user properties ($p$), rules ($r$), environment ($e$), time ($t$), perception ($P$), and natural language descriptions of features ($\chi$). The framework integrates all input information to simulate user decision-making in real scenarios, generating charging behavior decisions consistent with user charging behavior features. In particular, the statistical descriptions in $\chi$ provide soft guidance that helps the LLM calibrate its decisions toward realistic population-level patterns.

\subsubsection{Environment Tool Design}
To effectively implement the environment module in the EV charging behavior framework, the following tools have been developed.
% \paragraph{Energy Consumption Estimator}
% This tool estimates how much energy the EV consumes during trips, factoring in the vehicle’s efficiency and distance traveled.
\paragraph{Location Mapping Tool}
Using the Gaode Map API, this tool retrieves the latitude and longitude of a specified address. It provides accurate geographic data to locate the vehicle and nearby charging stations, aiding in route planning and charging station selection.
\paragraph{Distance Measurement Tool}
This tool calculates the distance between two points, utilizing addresses or coordinates. The Gaode Maps API powers this feature, helping to evaluate spatial distances affecting the EV’s charging behavior.
\paragraph{Charging Station Information Tool}
This tool gathers detailed information about nearby charging stations, including their location, charging power, and pricing. Such data is essential for optimizing charging station selection and ensuring cost-effective charging.

\section{Simulation results}
\subsection{EV Charging Behavior Simulation Process with LLM-based Agents}
This study introduces a dynamic modeling framework based on LLMs to simulate the charging behavior of EV users in dynamic environments. The same LLM is adopted throughout the framework to ensure semantic consistency. Specifically, the gpt-4o model is employed with a temperature setting of 0.1 to ensure deterministic and stable decision outputs\footnote{The source code, experimental datasets, and supplementary materials for this study are available at: \url{https://github.com/cgcui/LLM_Agent_EVCharging}.}. The framework integrates the user’s travel path, environmental perception, and behavioral needs to simulate real-time adjustments in travel and charging behaviors. By leveraging LLMs' advanced contextual reasoning and decision-making capabilities, the framework captures the complex interactions among psychological traits, environmental factors, and user decision processes. The simulation environment is constructed using Gaode Map, providing high-resolution geographic and charging infrastructure data. The simulation process is divided into initialization and daily simulation cycles, which collectively enable adaptive and realistic decision-making. Fig.~\ref{fig:Decision-Making} illustrates the architecture of the framework and the interactions between its components. Detailed experimental settings are provided in Appendix.

\subsubsection{Initialization Phase}
The initialization phase lays the foundation for simulating personalized agent behavior. User attributes are generated based on demographic data \cite{Sun2017}, capturing individual traits such as price sensitivity, range anxiety, and time sensitivity. These attributes are assigned to each agent, ensuring that simulations reflect diverse user behavior patterns. Each agent is initialized with modules for short-term memory, long-term memory, travel plans, and reflection, which collectively support adaptive decision-making. The simulation environment is constructed using Geographic Information System (GIS) data, integrating real-time information about dynamic travel conditions and charging station locations. These inputs allow the LLM to dynamically adjust the agent’s strategies based on user psychology and environmental constraints. 

\subsubsection{Daily Simulation Cycle}
The daily simulation cycle models the complete decision-making process, simulating how agents dynamically adapt their behavior based on real-time data, user attributes, and memory. This cycle involves three key steps: begin of day travel plan generation, decision-making loops, and end of day memory updates.

\paragraph{Daily Travel Plan Generation}
At the start of each simulation day, agents generate an initial travel plan based on user-specific attributes and predefined tasks. This initial plan acts as the baseline and is dynamically updated throughout the day to account for real-time changes.

\paragraph{Travel and Charging Decision Cycle}\hspace{0pt}\par
(i) Environmental Perception: The agent continuously collects and updates real-time environmental data, including travel information, traffic conditions, SoC, and charging stations. This data provides dynamic inputs for the agent's decision-making process. 
% By perceiving the environment, the agent understands traffic conditions and charging status and adjusts its travel and charging plans.

(ii) Travel Decision: The agent makes its travel decisions based on its planned route, real-time traffic conditions, and other dynamic environmental factors. Beyond destination distance, the agent continually updates its routing decisions based on evolving environmental data, maintaining real‑time accuracy and adaptability.

% In addition to considering the distance to the destination, the agent continuously updates its travel decisions in response to changes in environmental data, ensuring that its decisions remain accurate and adaptive to real-time conditions.
 
(iii) Charging Decision: Based on a comprehensive integration of cognitive, behavioral, and environmental information, the agent formulates its charging decision through a structured reasoning process. This process incorporates user-specific preferences, psychological traits, the current SoC, short-term memory and long-term memory. In parallel, the agent considers its ongoing travel plan and real-time environmental factors, including charging station availability, pricing, and spatial proximity. Furthermore, the agent is guided by statistical priors (\(\chi\)), explicitly defined as the empirical distributions of charging initial SoC and charging start times derived from real-world EV user behavior \cite{evcipa2022whitepaper}. All these components are jointly processed through the decision function \(\theta_L(p, r, P, e, t, \chi)\), producing the action \(a\) that determines whether to initiate charging and which charging station to select in the current context.
 
(iv) Short Memory Update: After each trip or charging session, agents store relevant data (e.gs., updated SoC, charging station usage, travel times) in short-term memory. This allows agents to refine subsequent decisions based on historical feedback, improving adaptability and precision across daily cycles.

\paragraph{Daily Reflection and Long-Term Memory Update}\hspace{0pt}\par
(i) Reflection:  At the end of each day, agents evaluate the effectiveness of their decisions through the reflection module. The reflection phase assesses whether decisions aligned with user-specific goals, such as minimizing travel time or reducing range anxiety, and identifies potential improvements.

(ii) Long-Term Memory Update: Insights from the reflection phase are stored in long-term memory to inform future decision-making. By combining real-time feedback with cumulative learning, the agent continuously refines its decision-making process to adapt to evolving environmental conditions and user behavior.

\subsection{Simulator Visualization} 

\begin{figure}[ht]
	\centering
	\includegraphics[width=0.5\textwidth]{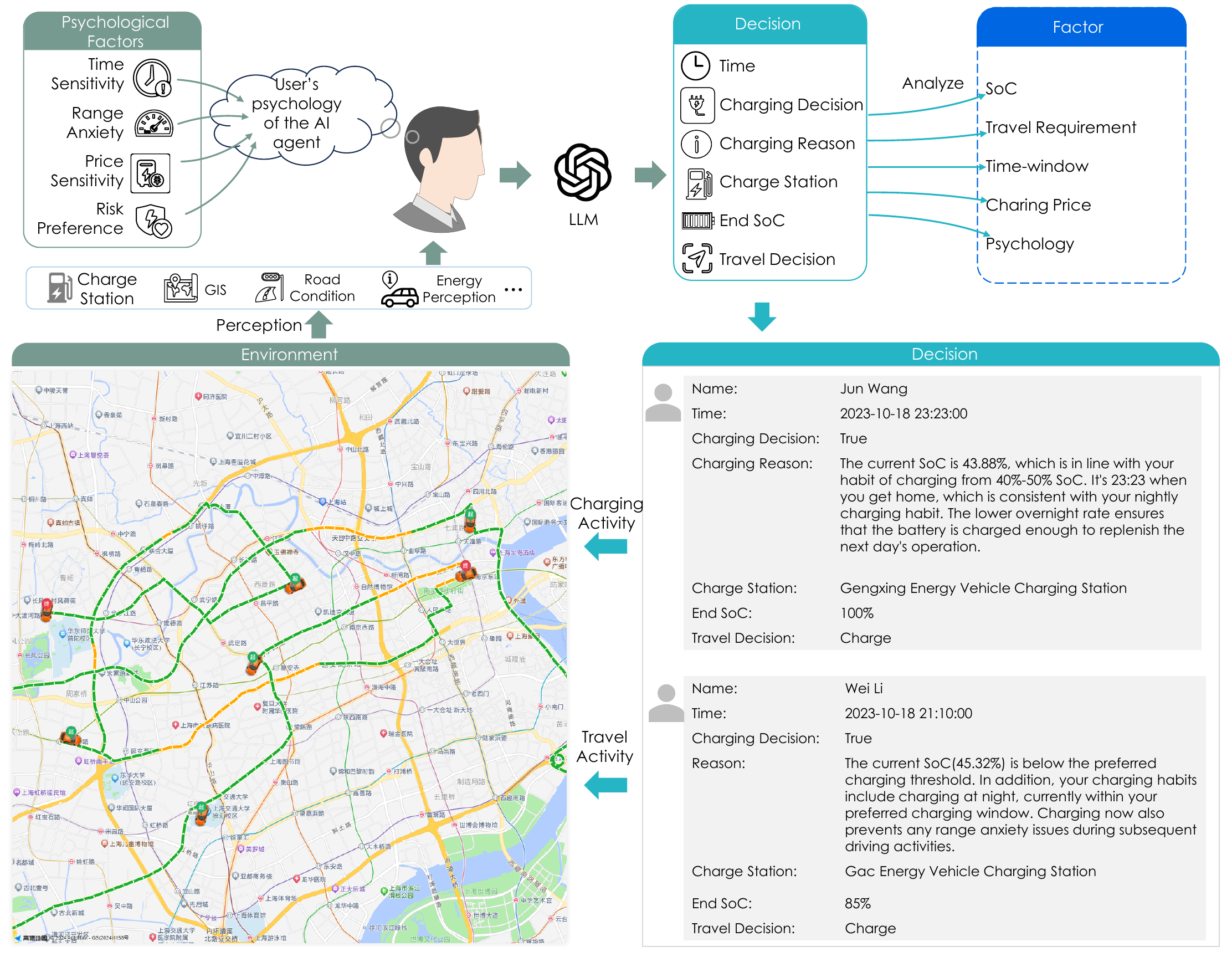}
	\caption{EV charging behavior simulation visualization. This figure illustrates the simulation framework integrating psychological factors like time and range sensitivity with decision-making on charging time, station selection, and capacity. The map shows charging station distribution and EV routes in Shanghai, while user logs highlight personalized behaviors. }
	\label{fig:EV_Charging_Behavior_simulation}
\end{figure}

Fig. \ref{fig:EV_Charging_Behavior_simulation} illustrates the dynamic simulation of EV charging behavior using the LLM-based agent framework, highlighting the interaction between environmental factors and the decision-making process. The left panel focuses on environmental inputs such as GIS data, travel times, and charging station availability, while the right panel displays the agent's decision-making process, which is influenced by both psychological factors and environmental data.

The left panel shows the dynamic environment in which the agent operates. It includes a GIS map that displays ev routes, travel information, and the spatial distribution of charging stations across the city. The real-time environmental data, such as the time required to travel between locations and the availability of nearby charging stations, are continuously updated, providing the agent with a real-time perception of the environment. This data helps the agent determine when and where to charge, depending on its current location and the time needed to reach the next destination.

The right panel captures the agent's decision-making process, which integrates psychological factors with real-time environmental inputs. The agent's charging decision is influenced by its SoC and the time to travel to the next location, as well as psychological traits such as range anxiety and time sensitivity.

\subsection{Simulation Results}
To investigate the intrinsic motivations and external triggers behind EV user charging behavior, a LLM-based agent framework to systematically analyze the complex interactions between psychological traits and dynamic environmental factors is employed. The utilization of EV taxi users in this study offers a unique advantage, as they represent a high-frequency charging scenario, providing rich data for analysis. To reflect the charging behavior of EV users more accurately, the behavior of 1,200 EV taxi drivers is simulated across three representative cities (Shanghai, Shenzhen, and Chengdu) in China over a one-week period. The simulation generated 14,279 charging decisions, each accompanied by a detailed description of the motivations driving these decisions. 

\subsubsection{Accuracy and Generalization of the Simulator}
To evaluate the effectiveness of the proposed LLM-based simulator, this study conducts two complementary experiments to address challenges in accuracy and generalization. The first experiment focuses on the simulator’s ability to replicate realistic EV user charging behaviors, validating its reliability in capturing core decision-making patterns. The second experiment tests the simulator's adaptability across different urban environments, verifying its generalization capability and robustness in diverse scenarios.

\begin{figure*}[h]
	\centering
	\includegraphics[width=1.0\linewidth]{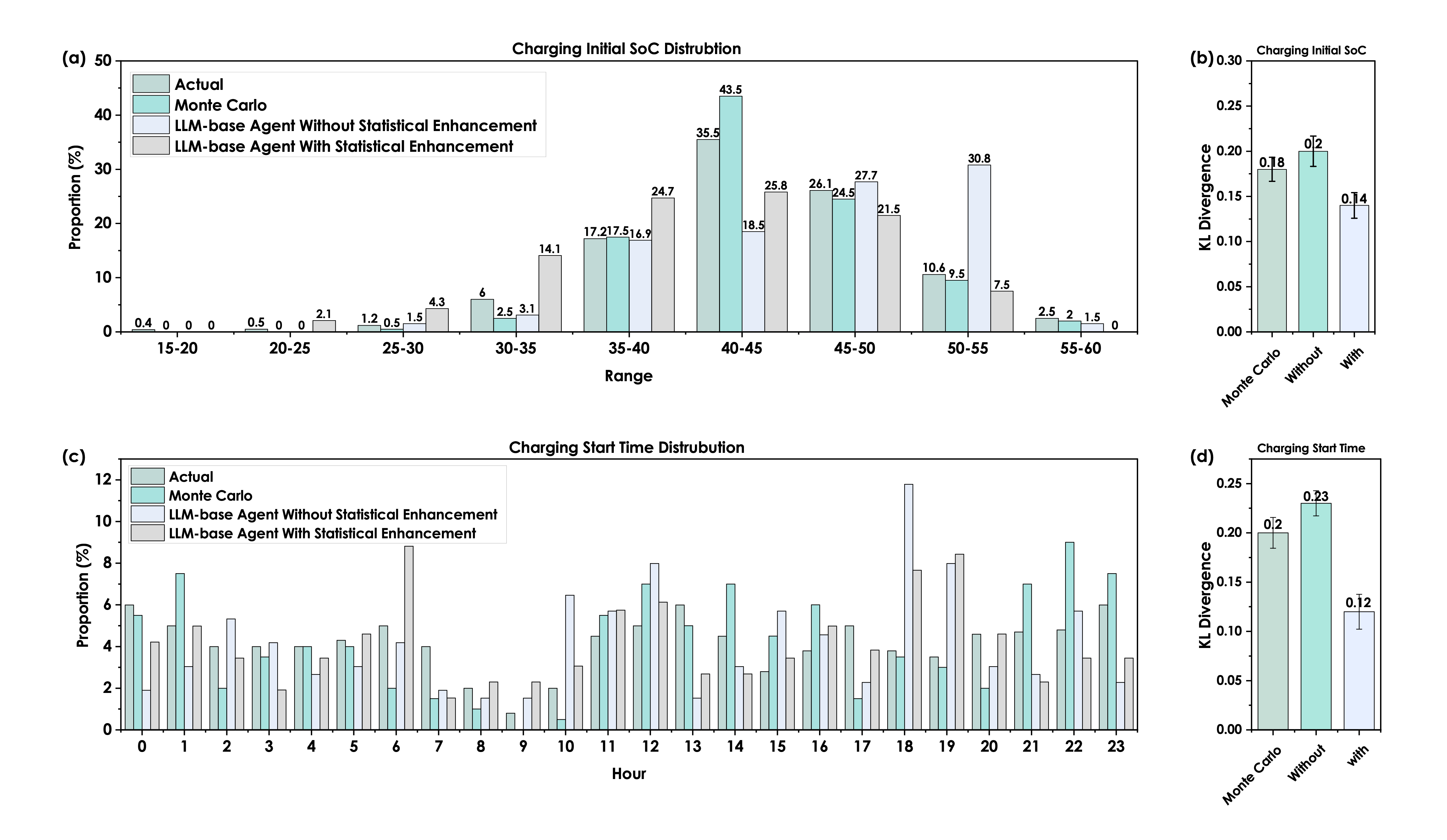}
	\caption{(a) Charging initial SoC distribution, (b) KL divergence of charging initial SoC distributions comparing Monte Carlo simulation, LLM-based agent without statistical enhancement, and LLM-based agent with statistical enhancement, (c) Charging start time distribution, (d) KL divergence of charging start time distributions comparing Monte Carlo simulation, LLM-based agent without statistical enhancement, and LLM-based agent with statistical enhancement.}
	\label{fig:Accuracy_simulation}
\end{figure*}

\paragraph{Accuracy of the Simulation}
This study employs two key metrics—charging initial SoC distribution and charging start time distribution—from real-world data to assess the accuracy of the simulator in reproducing user charging behavior. These metrics provide reliable benchmarks for evaluating the simulator’s ability to reflect realistic decision-making processes. A comparative analysis with conventional Monte Carlo simulation \cite{kamana2024driving} methods demonstrates that the proposed LLM-based approach more accurately captures the temporal patterns and heterogeneity of user behavior, thereby providing stronger behavioral fidelity and interpretability in dynamic charging scenarios.

Fig.\ref{fig:Accuracy_simulation}(a) and Fig.\ref{fig:Accuracy_simulation}(c) compare the distributions of charging initial SoC and charging start time, respectively, across four scenarios: real-world data, Monte Carlo simulation, a baseline LLM-based agent, and a statistically enhanced LLM-based agent. For both metrics, the baseline agent shows noticeable deviations from real-world behavior—overestimating high SoC levels and evening peak charging times. The Monte Carlo simulation captures partial patterns but suffers from inconsistencies and volatility. In contrast, the enhanced agent exhibits the closest alignment with empirical distributions, effectively replicating key behavioral trends across both SoC levels and temporal charging behavior.

To quantitatively evaluate the similarity between simulated and real-world distributions, Kullback–Leibler (KL) divergence is employed, as shown in Fig.\ref{fig:Accuracy_simulation}(b) and Fig.\ref{fig:Accuracy_simulation}(d). For the initial SoC distribution, KL divergence values are 0.178 for the Monte Carlo simulation, 0.200 for the baseline LLM agent, and significantly lower at 0.140 for the enhanced agent. Similarly, for the charging start time distribution, the enhanced agent achieves a KL divergence of 0.123, outperforming both the Monte Carlo (0.200) and baseline LLM agent (0.228). These results underscore the improved accuracy and robustness of the statistically enhanced LLM agent in capturing fine-grained charging behaviors.

\begin{figure}[ht]
	\centering
	\includegraphics[width=1\columnwidth]{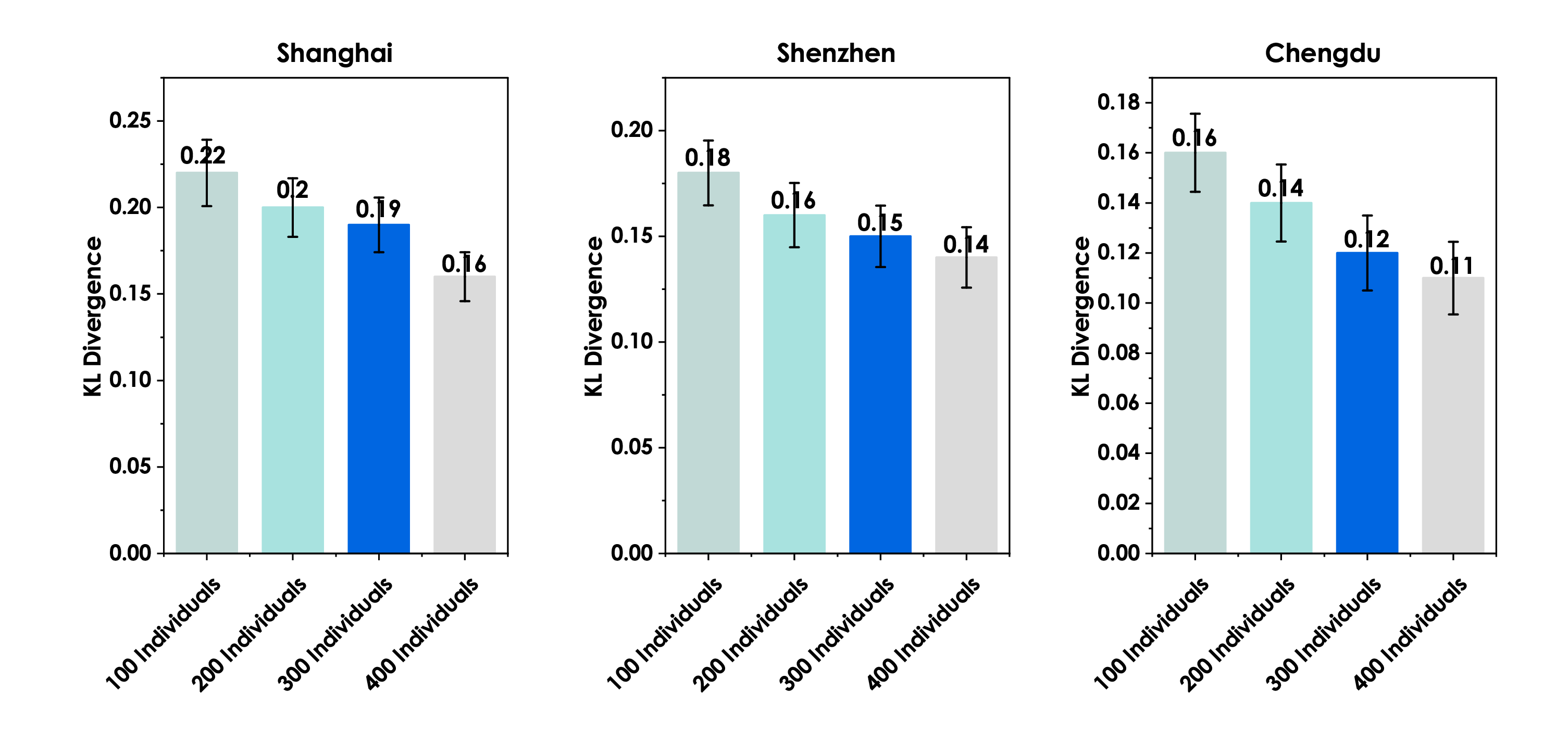}
	\caption{Comparisons of EV charging behavior distributions in different urban environments.}
	\label{fig:different_city_simulation}
\end{figure}

\paragraph{Generalization of the Simulation}
The generalization capability of the proposed method is evaluated by simulating EV user charging behaviors across different cities and comparing the results with actual data. Fig.~\ref{fig:different_city_simulation} shows the probability distributions of simulated charging behaviors and actual data in urban environments, including Shanghai, Shenzhen, and Chengdu. The KL divergences  between simulated results and actual data in these cities decrease as the increase in population, respectively, indicating that the method effectively replicates actual distributions. For example, in Shanghai, as the population increases, the KL divergence decreases from 0.22 to 0.16.

The findings of this study demonstrate the robustness and versatility of the proposed LLM-based agent framework in simulating EV user charging behaviors. By incorporating statistical knowledge enhancement, the framework achieves significant improvements in accuracy, effectively replicating both SoC and charging time distributions. Moreover, its strong generalization capability across diverse urban environments highlights its adaptability and potential for practical applications.

\subsubsection{Contextual Learning for Charging Decision Analysis}
A contextual learning framework for quantitatively analyzing the factors influencing EV charging decisions is proposed in this paper. The framework models the relationship between decision texts generated by LLMs and user behavior triggers, shedding light on the impact of psychological traits and environmental factors on charging decisions. By integrating semantic analysis, mathematical representations, and contextual learning, this approach  addresses the complexity of dynamic charging environments.

\paragraph{Role of Behavioral Triggers in Charging Decisions}
Behavioral triggers are key variables influencing EV users' decisions, including intrinsic psychological traits and external environmental factors. These triggers help identify the motivations behind user behavior and analyze how users respond to dynamic conditions. By incorporating these triggers, the framework links decision texts to quantifiable factors, providing a structured analytical approach to model user behavior.

\paragraph{Mathematical Representation of Charging Decision Triggers}

To capture the impact of behavioral triggers on charging decisions, the framework uses a multi-dimensional binary vector representation. This method concisely describes the relevance of triggers and mathematically maps the motivations behind user decisions.

Behavioral triggers are represented as a multi-dimensional binary vector:
$\mathbf{z} = [z_1, z_2, \dots, z_k]^{\top}$
where \(k\) corresponds to the total number of trigger dimensions and \(z_j \in \{0, 1\}\) indicates whether the \(j\)-th trigger dimension is related to the decision text \(T_{\text{decision}}\), defined as:
\begin{equation}
z_j =
\begin{cases}
1, & \text{if the trigger dimension } j \text{ is relevant to } T_\text{decision} \\
0, & \text{otherwise}
\end{cases}
\end{equation}

This structured representation enables a compact and rigorous description of the relationship between decision texts and behavioral triggers, facilitating a mathematical mapping of user behavior to its underlying drivers.

\paragraph{Contextual Learning for Decision Analysis}

The contextual learning framework evaluates the semantic association between decision texts and behavioral triggers, identifying the key factors behind decisions. The input is constructed by combining predefined query templates with the decision text, providing the LLM with the necessary context:
\begin{equation}
X_{\text{ICL}} = D_{\text{prompt}} \parallel T_{\text{decision}}
\end{equation}
where \(D_{\text{prompt}}\) is a set of query templates used to provide context for the LLM. The query templates are formally defined as:
\begin{equation}
D_{\text{prompt}} = \{(x_j, z_j) \mid x_j \in X,\ z_j \in \{0, 1\},\ j = 1, \dots, k\}
\end{equation}

\paragraph{Decision Text Analysis}

Given the input prompt \(X_{ICL}\) and the label set \(\mathcal{Y} = \{y_1, y_2, \dots, y_k\}\), the LLM’s response function \(\theta_{LLM}\) generates binary outputs for each trigger dimension. The charging decision vector \(\mathbf{z}\) is computed as:
\begin{equation}
z_j = \theta_{\text{LLM}}(X_{\text{ICL}}, y_j), \quad j = 1, \dots, k
\end{equation}
where \(\theta_{LLM}\) determines the association between \(T_{\text{decision}}\) and the trigger dimension \(y_j\). If \(T_{\text{decision}}\) is relevant to \(y_j\), then \(z_j = 1\); otherwise, \(z_j = 0\). The final charging decision vector \(\mathbf{z}\) is derived by aggregating all the outputs for each trigger dimension.

\subsubsection{Analysis of Environmental Factors}
In what follows, the impact of environmental factors on the charging decisions of EV users is investigated, and the mechanisms of distance to charging stations and charging prices are discussed. The dataset utilized in this study comprises 5775 charging decisions made by 500 EV users over the span of one week. The analysis result is presented as follows.

\begin{figure}[ht]
\centering
\includegraphics[width=1.0\columnwidth]{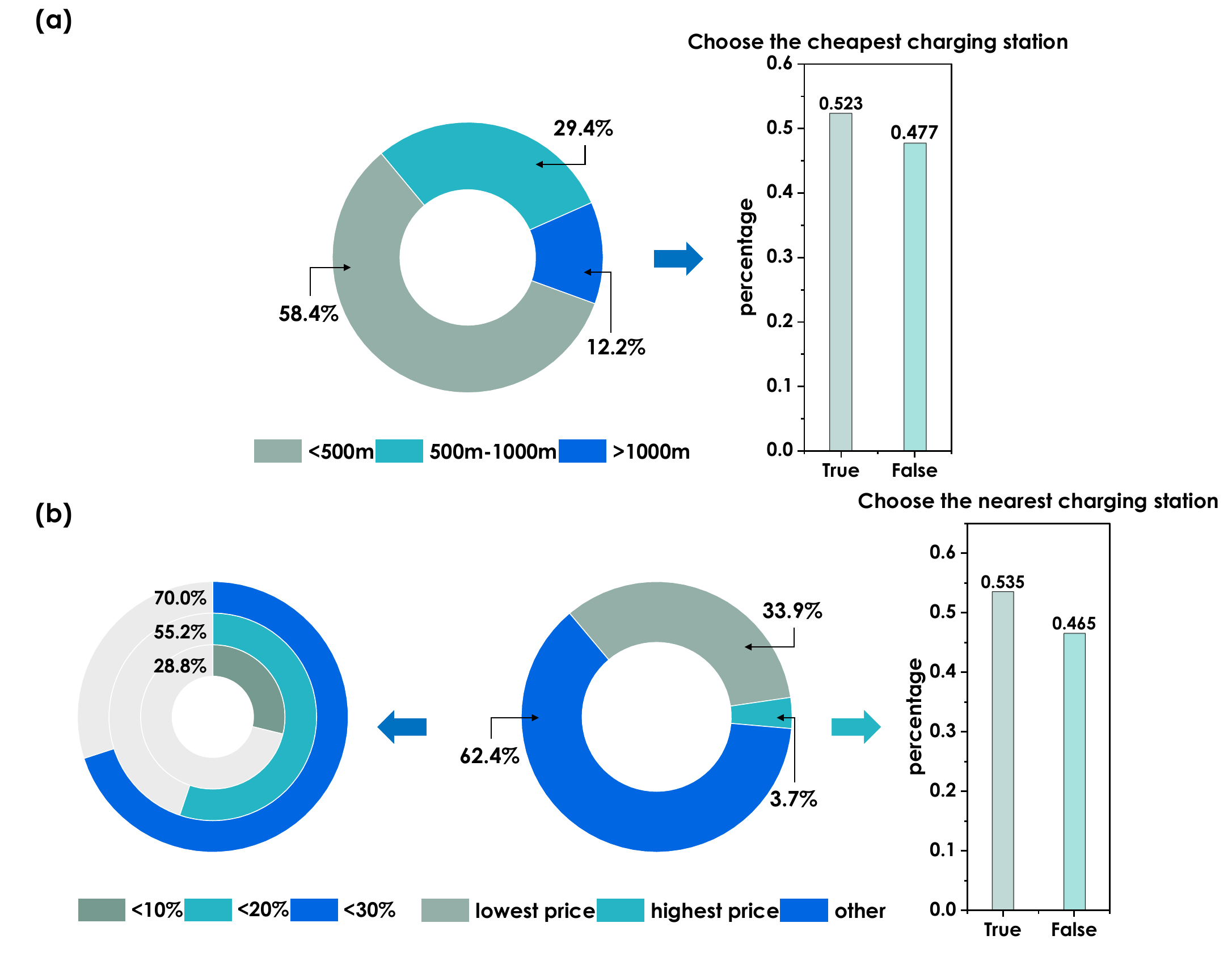}
\caption{The influence of distance on charging decision (a), the influence of price on charging decision (b).}
\label{fig:Enviroment_factor}
\end{figure}

As shown in Fig.~\ref{fig:Enviroment_factor}(a),  58.4\% of charging decisions involve selecting stations within 500 meters, underscoring the importance of proximity and reflecting users’ clear preference for geographically nearby options. Additionally, 12.2\% of decisions involve stations located more than 1000 meters away. Further analysis reveals that 52.3\% of these long-distance selections are for the lowest-priced stations, suggesting that charging price is more likely to be considered a decisive factor in decision-making when charging stations are located farther away.

With regard to charging price, charging stations offering moderate prices are selected in 62.4\% of charging decisions, as shown in Fig.~\ref{fig:Enviroment_factor}(b). Further analysis indicates that 70\% of these decisions involve charging price within 30\% of the lowest available rate, implying that users are highly price-sensitive in their charging decisions, though not an exclusive preference for the lowest price. The significance of price in decision-making is further validated by the observation that the lowest-priced charging stations are selected in 33.9\% of charging decisions. Additionally, charging stations with the highest price are chosen in 3.7\% of charging decisions. It is determined that 53.5\% of the decisions favored the nearest charging stations, as further analysis indicated. The result shows that in certain cases, regional convenience may surpass cost factors.

\subsubsection{Analysis of Psychological Factors}
\begin{figure*}[ht]
\centering
\includegraphics[width=1.0\linewidth]{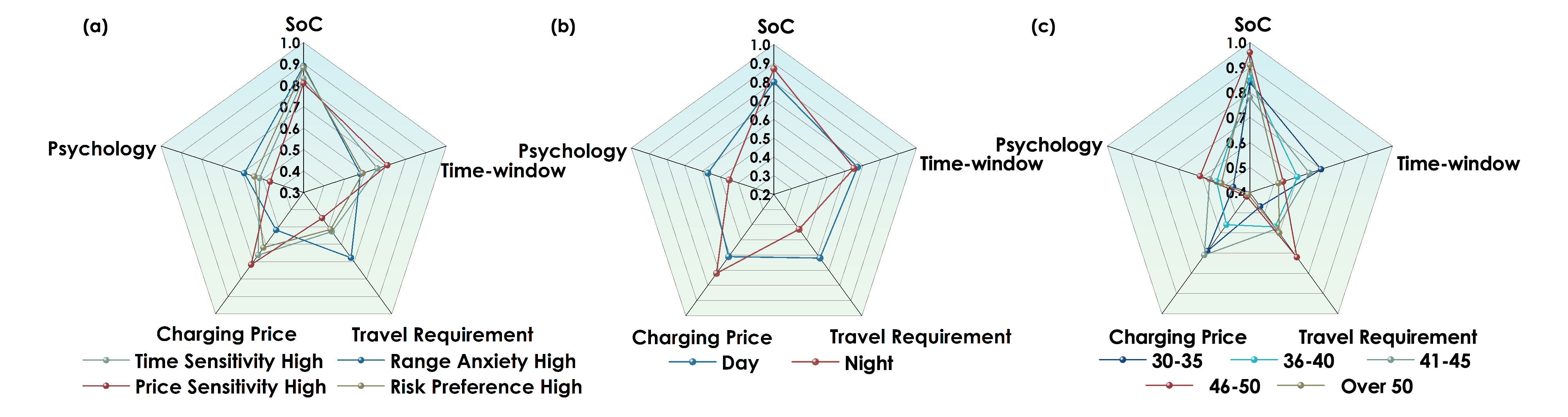}
\caption{Influencing factors of charging decisions for different psychological groups (a), day shift and night shift groups (b), and different age groups (c).}
\label{fig:Experimental analysis result}
\end{figure*}

This study employs a contextual learning approach combined with LLM-based agent simulations to explore the causal relationships between psychological traits and charging decisions. The dataset comprises 14,279 charging decisions generated by 1,200 EV users over one week. The psychological traits include time sensitivity \cite{alsabbagh2020distributed}, range anxiety \cite{rainieri2023psychological}, price sensitivity, and risk preference, each classified into high, medium, and low. The causal relationships between these traits and five key decision-making factors (time-window, charging price, travel requirement, SoC, and psychology) are analyzed, with the results shown in Fig.~\ref{fig:Experimental analysis result}(a). The analysis is summarized as follows.

Users with high time sensitivity place strong emphasis on SoC (83.2\%) and time-window (66.2\%), while the influence of travel requirement is lower (52.6\%). It is indicated that individuals with high time sensitivity favor short-term rewards over future benefits. For users with high range anxiety, both SoC (89.0\%) and travel requirement (67.7\%) play key roles, whereas time-window and price have less impact. These users tend to charge preemptively to reduce uncertainty and maintain security.

Users with high price sensitivity exhibit a strong preference for minimizing charging costs, especially by taking advantage of low-cost charging periods. While 81.0\% of charging decisions are influenced by SoC, charging price emerges as the primary factor, affecting 71.1\% of decisions. Travel requirement has the least impact at 44.8\%, confirming that price-sensitive users are primarily driven by the goal of minimizing costs, rather than optimizing travel range. In contrast, users with high risk preference adopt a more balanced strategy, considering both SoC (88.1\%) and price (61.8\%), aiming to maintain sufficient charge while avoiding unnecessary expense.

\subsubsection{Heterogeneity in User Behavior}
To investigate the heterogeneity in user charging behaviors, a comprehensive analysis of charging data based on the demographic attributes of 1200 simulated users is conducted. This study primarily analyzes the charging decisions behaviors among individuals with different working schedules and across various age groups, as illustrated in the results below.

Fig.~\ref{fig:Experimental analysis result}(b) illustrates the differences in charging decision triggers between day-shift and night-shift users. The analysis reveals that day-shift users primarily prioritize time-window preferences, often charging when their range is insufficient or when specific travel needs arise, exhibiting lower sensitivity to charging price. In contrast, night-shift users are more inclined to consider economic factors, leveraging lower electricity prices during nighttime, and demonstrating a higher focus on cost efficiency. 

This study further investigates the influence of age groups on user charging behavior, as illustrated in Fig.~\ref{fig:Experimental analysis result}(c). The findings reveals that individuals aged 30–35 and 36–40 display comparable behavioral patterns, with their charging decisions predominantly influenced by time-window preferences to enhance travel efficiency. In contrast, users aged 41–45 exhibit a stronger sensitivity to price, with charging decisions being more influenced by cost considerations, highlighting a greater focus on economic factors. For users aged 46–50 and those aged 50 and above, charging decisions are predominantly influenced by travel requirements, with the main objective being to ensure that their mobility needs are met.

\subsubsection{Integrated Effects Analysis of Environmental and Psychological Factors}
In what follows, further analysis is conducted to investigate how environmental factors interact with psychological traits, particularly price sensitivity and range anxiety, to jointly influence the charging decisions of EV users. A detailed analysis is conducted based on the demographic profiles of 500 simulated users, with the results summarized below.

\begin{figure}[ht]
\centering
\includegraphics[width=1.0\linewidth]{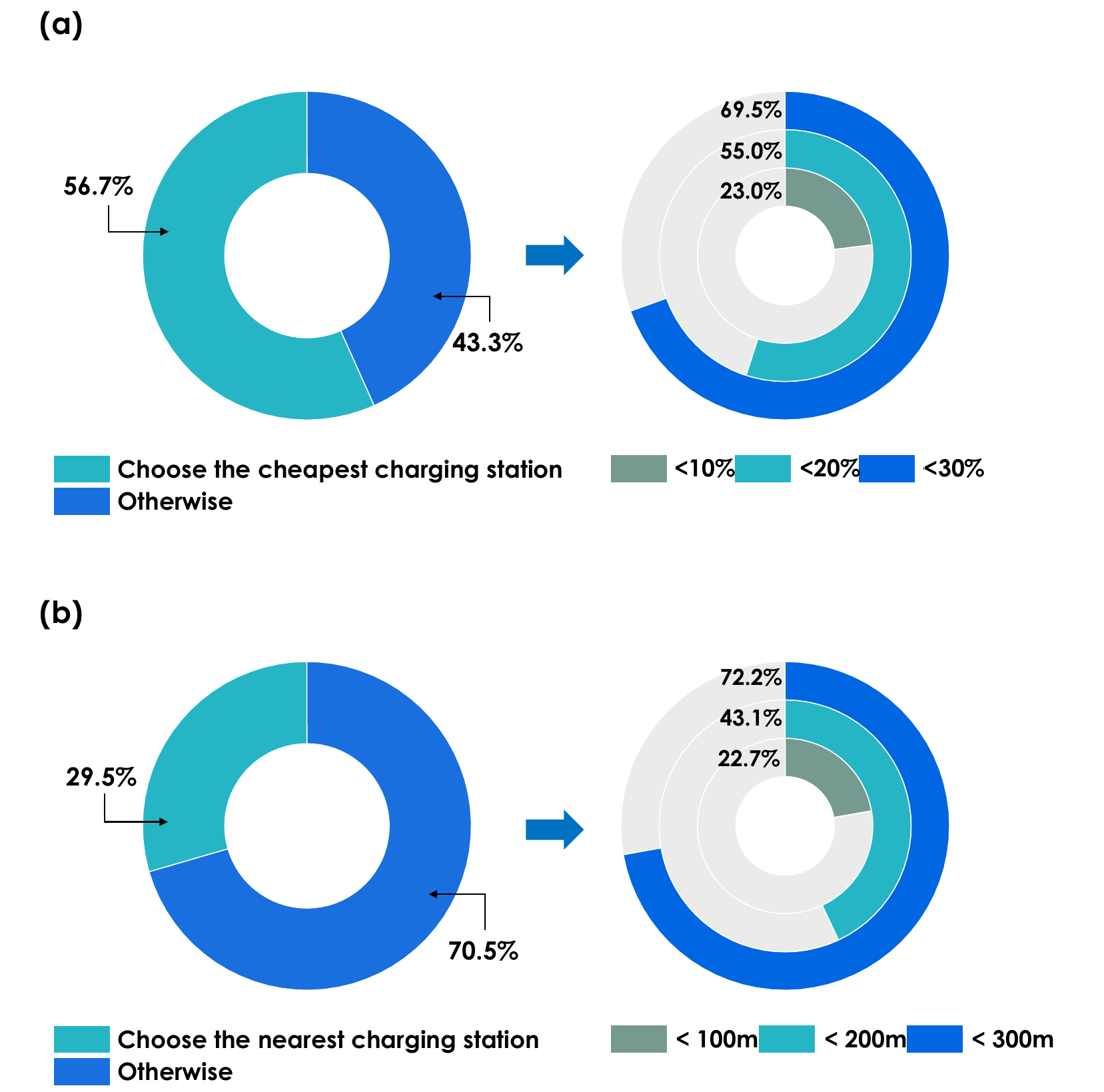}
\caption{The impact of price on charging decisions of high price-sensitive users (a), the impact of distance on charging decisions of high range anxiety users (b).}
\label{fig:Enviroment_factor_Psychology_factor}
\end{figure}

According to the data in Fig.~\ref{fig:Enviroment_factor_Psychology_factor}(a), 56.7\% of charging decisions by highly price-sensitive users involve selecting the lowest-priced stations, highlighting the strong influence of cost in this subgroup. However, 43.3\% of decisions do not involve the lowest-priced option. Further analysis shows that 69.5\% of these cases involve charging price within 30\% of the lowest available rate. These results indicate that although price is a key factor, other considerations are also integrated into the decision-making process, such as proximity and charging power.

As illustrated in Fig.~\ref{fig:Enviroment_factor_Psychology_factor}(b), while 29.5\% of charging decisions made by users with high range anxiety favor the nearest charging station, 70.5\% select alternative options. According to further analysis, 72.2\% of these alternative decisions involve selecting charging stations located within 300 meters of the nearest option. Although range anxiety affects decision-making, users are also found to incorporate additional factors, such as charging prices and charging power, into their final choices, as shown by these findings.

The findings from this analysis highlight the critical role of both environmental and psychological factors in shaping EV users' charging decisions. By simulating users with diverse psychological profiles, the LLM-based agent accurately captures the influence of different psychological dimensions in the decision-making process. Moreover, this method reveals the complex interactions between intrinsic motivations and decision triggers, providing profound insights into user-specific charging behavior patterns and establishing a robust theoretical foundation for behavior modeling and optimizing charging infrastructure design.

\subsection{Discussion on Computational Efficiency and Scalability}
The computational overhead of the proposed framework is mainly dominated by LLM inference and environment-tool querying. In this work, a decision cycle is defined to include one charging decision, one charging-station query, and one route/traffic planning operation to support travel and charging decision-making as well as charging-station selection. Each decision cycle triggers a fixed number of LLM calls under structured prompts to generate the charging decision and reason; consequently, the total number of LLM calls grows proportionally with the total number of decision cycles. Beyond these decision-dependent calls, the framework requires a small number of initialization calls to generate default user and context information at the beginning of the simulation (five LLM calls in this work).

Since the number of decisions per day is not fixed and varies with daily mobility patterns and triggering events, the computational budget is governed by the total number of decisions generated during the simulation rather than the nominal number of simulated days. Consider a simulation with $N$ agents (users) over a horizon of $T$ days. Let $K_{i,t}$ denote the number of decisions generated by agent $i$ on day $t$, and let $K_{\mathrm{tot}}=\sum_{i=1}^{N}\sum_{t=1}^{T}K_{i,t}$ denote the total number of decisions. The overall runtime scales approximately linearly with $K_{\mathrm{tot}}$, and can be summarized as $O(K_{\mathrm{tot}})$ up to constant factors determined by the fixed LLM-call budget per decision and the per-decision environment operations (one charging-station query and one route planning). 

The framework is scalable in practice because agents are conditionally independent given the environment, enabling parallel execution across agents and across days. Moreover, caching frequently accessed station-query results and commonly used route segments can reduce redundant tool calls in shared spatial contexts, and batched execution further improves throughput when scaling to large agent populations. These characteristics support the use of the proposed framework in planning-oriented studies that require large-scale generation and analysis of heterogeneous charging behaviors.

\section{Conclusion}
This study introduces a novel framework for analyzing the charging behavior of EV users, integrating psychological traits and situational factors into a unified simulation architecture. Simulation results across multiple cities demonstrate that the proposed framework achieves high accuracy in replicating EV users’ charging behaviors and exhibits good generalization performance across different urban environments. Further analyses reveal the joint effects of environmental and psychological factors on charging decisions, offering insights into the underlying behavioral mechanisms and population heterogeneity, and providing support for charging infrastructure planning and charging load assessment.

Despite the proposed framework offering valuable insights, it still presents several limitations. The current instantiation and validation mainly focus on EV taxi drivers and do not cover a broader population of private EV owners. The simulation horizon is one week, which reflects typical weekly dynamics but does not capture longer term variability such as holidays, special events, or seasonal changes. The analyses in this paper are conducted based on simulation generated data. During decision generation, external statistical priors and environmental information are incorporated, and the simulator outputs are benchmarked against external empirical statistics on representative charging features. Based on this benchmarked consistency, further analysis is conducted to examine how situational triggers and psychological traits influence decision tendencies within the simulated scenarios. As a result, the findings may still involve deviations and uncertainty when extrapolated to more complex real world settings and broader user populations. Furthermore, using the same LLM across multiple stages may reduce the independence between generation and analysis, thereby introducing systematic bias in the trigger interpretation. Prompt engineering and RAG are adopted to reduce hallucination risks, but occasional hallucinations or inconsistencies may still occur due to inherent limitations of generative models.

Future work will extend the framework to private EV owners and improve the representation of broader EV user populations. The simulation horizon will be extended beyond one week to incorporate holidays, special events, and seasonal factors, and richer dynamic variables such as fine grained traffic and weather information, user mobility, and charging accessibility will be integrated. Adaptive modeling and reinforcement learning techniques will also be explored to enhance robustness under evolving environments and to support more practical planning evaluation and operational analysis.

\section*{Appendix}
\subsection{Simulation Setup}
In this study, a real geographic environment is utilized as the simulation model to replicate the charging behavior of 100 taxi drivers over a seven-day period. Each EV starts with a battery energy of 42.755 kWh (total capacity of 50.3 kWh). The 100 taxi drivers are divided into two shifts, day and night, each consisting of 50 drivers. A simulation is conducted using GPT-4o, with a temperature setting of 0.1. The Amap API is used to obtain information such as calculating the distance between two locations and identifying nearby charging stations. 
\paragraph{Property Generation}Based on whether the taxi driver is assigned to the day or night shift, the LLM uses prompts and RAG to generate personal information, psychological characteristics, EV details, and charging habits.
\paragraph{Rule Generation} The rule generation process involves three phases: planning, memory storage, and reflection, which guide the agent’s decision-making process. Initially, the LLM creates a preliminary plan based on the daily data of taxi drivers in a city and refines it according to trip-specific needs, resulting in a final plan. Then, decision information is stored in memory: short-term memory holds travel plans and decisions from the past three days, while long-term memory contains data from the last seven days. Finally, at 12:00 each day, the LLM conducts a daily reflection, reviewing whether the day’s decisions met user expectations, assessing satisfaction with charging choices, and evaluating their cost-effectiveness.

\paragraph{Environmental Perception}The Amap API is used to obtain information such as calculating the distance between two locations and identifying nearby charging stations. Charging station information is matched with the database, assuming that all stations are available.

\paragraph{Behavior Simulation} At the end of each event, the vehicle's energy consumption is calculated, and a charging decision is made. Relevant information is provided to the LLM to make decisions that align with the user profile, and these decisions are stored in the database.

\begin{table}[htbp]
\centering
\caption{Experiment Setup for Simulating EV Charging Behavior}
\label{table:experiment_setup}
\renewcommand{\arraystretch}{1.4} % Adjust the row height for better readability
\resizebox{\linewidth}{!}{
\begin{tabular}{|p{4.5cm}|p{7.5cm}|}
\hline
\textbf{Parameter}  & \textbf{Details} \\ \hline
\textbf{Environment} & Shanghai, China \\ \hline
\textbf{Duration} & 7 days \\ \hline
\textbf{Role} & Taxi Drivers \\ \hline
\textbf{Number of people} & 100 (50 daytime drivers, 50 nighttime drivers) \\ \hline
\textbf{Initial SoC} & 42.755 kWh (Total capacity: 50.3 kWh) \\ \hline
\textbf{Vehicle type} & Electric Vehicles \\ \hline
\textbf{LLM} & GPT-4o \\ \hline
\textbf{Temperature} & 0.1 \\ \hline
\textbf{Property} & Generates driver profiles. \\ \hline
\textbf{Perception} & Use the Amap API to get environmental information. \\ \hline
\textbf{Rule} & Generates work plans, memory and daily reflections. \\ \hline
\textbf{Action} & Simulates charging decisions. \\ \hline
\end{tabular}
}
\end{table}

\subsection{Impact of Parameter in LLMs on Simulator} 
In order to verify the effectiveness of the proposed method, this research thoroughly examines the model’s performance from two key angles: the influence of critical parameters on model accuracy and the model’s ability to generalize across various conditions. The primary experimental findings and analyses are summarized as follows.

\begin{figure}[ht]
\centering
\includegraphics[width=0.95\columnwidth]{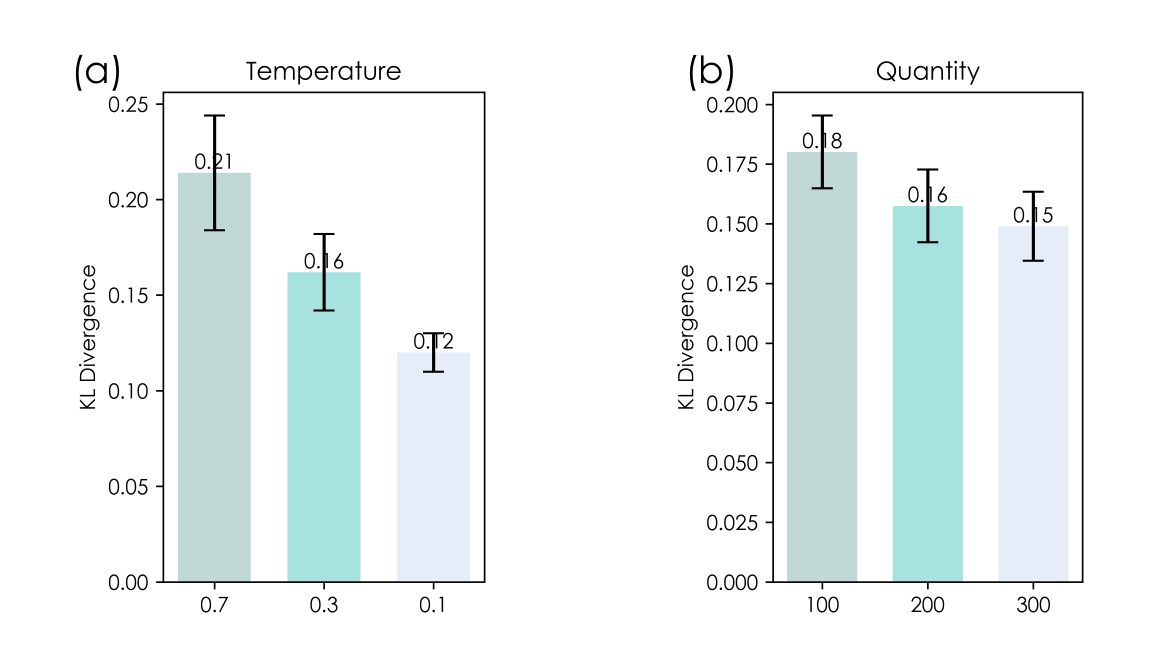}
\caption{The impact of different temperatures and varying quantities.}
\label{fig:influence_simulation}
\end{figure}

\paragraph{Impact of Temperature Parameter in LLMs on Simulator}
Fig.~\ref{fig:influence_simulation}(a) illustrates how the temperature parameter in a large language model affects the accuracy of the initial SOC in charging behavior simulation. The results indicate that at lower temperatures (e.g., 0.1), the model’s generated behavior is more concentrated, showing greater determinism. Conversely, at higher temperatures (e.g., 0.7), the randomness in behavior increases, resulting in a higher KL divergence and a more dispersed behavior distribution. This suggests that the temperature parameter can effectively modulate the balance between randomness and determinism, allowing customization based on application needs. However, an excessively high temperature may lead to overly random behavior, potentially reducing the model’s ability to accurately reflect actual user preferences. Consequently, it is essential to carefully calibrate the temperature parameter in practical applications to maintain model stability and ensure a realistic behavior distribution. 

% \begin{figure*}[ht]
% \centering
% \includegraphics[width=1.0\textwidth]{Simulation of charging behavior at different temperatures.pdf}
% \caption{Simulation of Charging Behavior at Different Temperatures}
% \label{fig:temmperatures_simulation}
% \end{figure*}

\paragraph{Influence of Electric Vehicle Quantity on Simulator}
Fig.~\ref{fig:influence_simulation}(b) shows how different numbers of EVs affect the accuracy of start-time simulation in charging behavior. The results reveal that as the EV count rises (for example, from 20 to 100), the behavior distribution generated by the model becomes more concentrated, and the alignment with the original distribution improves markedly. This relationship is confirmed by the changes in KL divergence: at 20 EVs, the KL divergence stands at 0.21, indicating a significant mismatch with the original distribution; as the EV count increases to 50, the KL divergence decreases to 0.13; and with 100 EVs, it further drops to 0.11. These findings demonstrate that a higher EV count enhances the model’s ability to accurately replicate charging behavior distribution, bringing it closer to real-world data.

In more detail, with a smaller EV count, the model finds it challenging to create a stable behavior distribution, resulting in a larger deviation from the original data. However, as the EV count increases, the model’s behavior distribution gradually becomes more focused, resulting in a higher level of alignment with the actual distribution. This trend indicates that increasing the EV count not only improves simulation accuracy but also enhances the model’s applicability in areas such as demand forecasting and load management.

\subsection{An Example of EV User Decision Making}
\begin{figure}[ht]
\centering
\includegraphics[width=0.95\columnwidth]{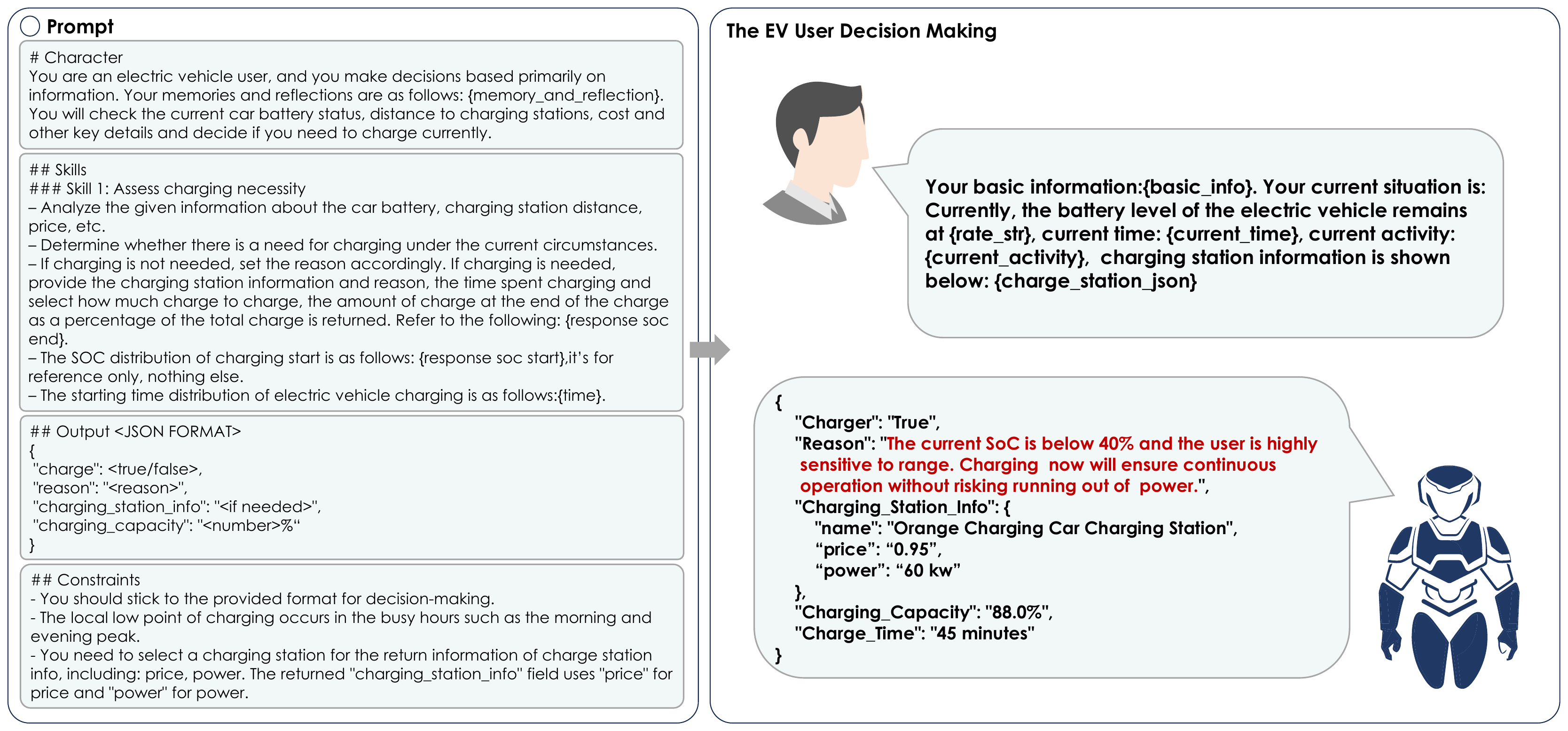}
\caption{Case studies on EV user charging decisions. The left side indicates prompt words, while the right side represents the actual dialogue process.}
\label{fig:prompt_simulation}
\end{figure}

In this section, an illustrative example  of the decision-making process is presented undertaken by an EV user.  As shown in {Fig. \ref{fig:prompt_simulation}}, the scenario demonstrates a structured approach where the user evaluates several factors, including the current SoC, distance to charging stations, charging costs, and other key considerations, to determine the necessity of charging.

The user's initial situation is characterized by a SoC of {rate\_str}, the current timestamp being {current\_time}, and the user engaged in {current\_activity}.  Information regarding available charging stations is summarized in Table 1 as follows: {charge\_station\_json}.  To assess the need for charging, the user systematically examines data on the remaining SoC, the proximity of charging stations, and the associated costs.  If charging is deemed unnecessary, the rationale for this decision is documented. Conversely, if charging is required, the user selects an appropriate charging station based on the provided parameters (e.g., station name, price per kWh, and power capacity) and determines the amount of charge needed as well as the estimated charging time.

For example, in the illustrated scenario, the user’s SoC is below 40\%, indicating a high risk of running out of power.  As a result, the user opts to charge immediately to ensure uninterrupted operation.  The selected charging station, ``Orange Charging Car Charging Station" offers a competitive price of CNY 0.95 per kWh and a power capacity of 60 kW.  The estimated charging requirement is 88.0\%, with a projected charging time of 45 minutes.  This example highlights the importance of a systematic and data-driven decision-making process for EV users, ensuring optimized charging strategies that balance convenience, cost, and operational efficiency.

\subsection{An Example of EV User Charging Decision Analysis}
In this section, an example analysis of the factors influencing EV users' charging decisions is presented. As shown in the Fig. \ref{fig:CoT Prompt} below, this scenario employs a structured approach, systematically analyzing each factor to elucidate their specific mechanisms of influence on user decision-making.

\begin{figure}[ht]
\centering
\includegraphics[width=0.95\columnwidth]{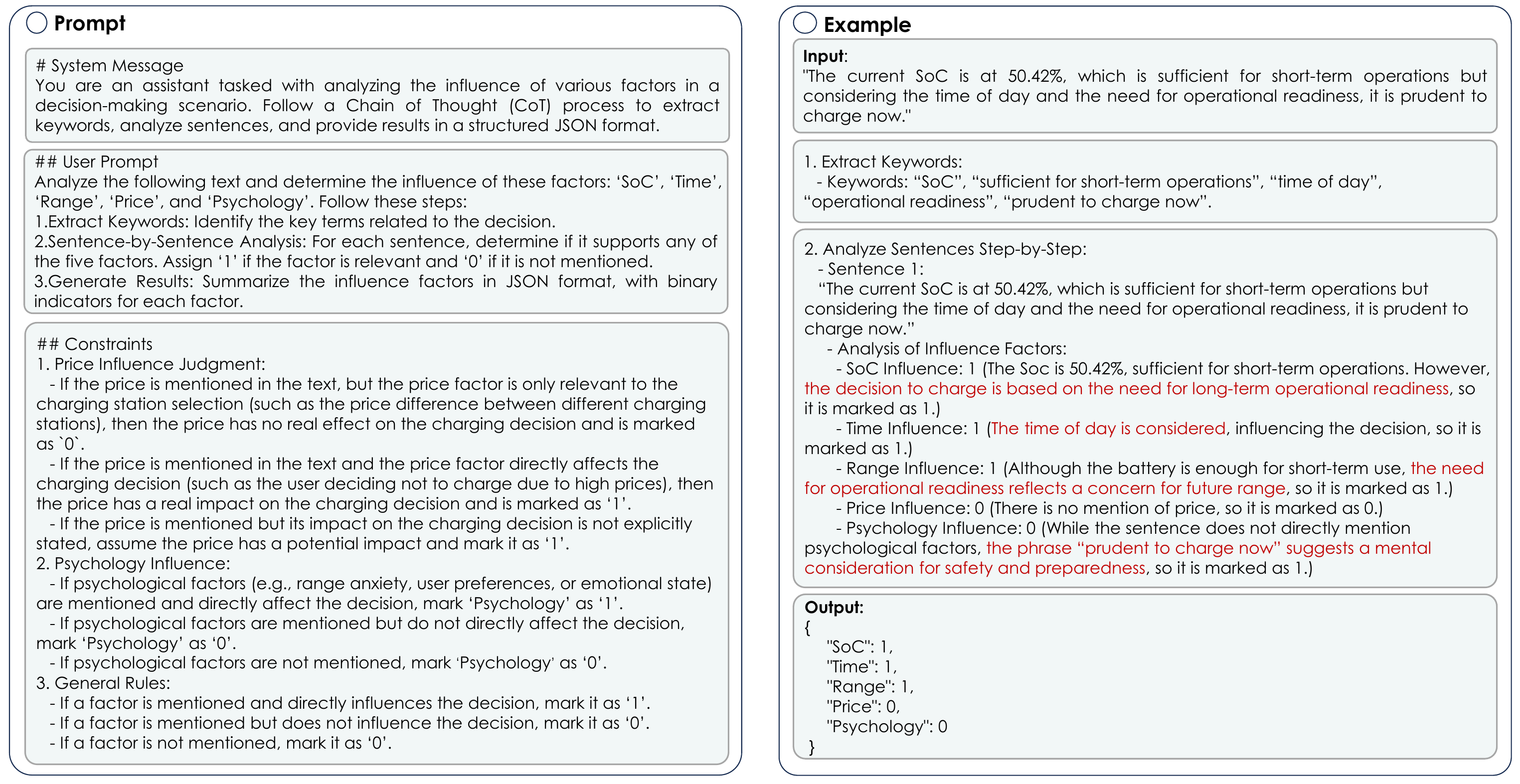}
\caption{Case studies on EV user charging decisions analysis. The left side indicates prompt words, while the right side represents the actual dialogue process.}
\label{fig:CoT Prompt}
\end{figure}

In the illustrated scenario, the rationale behind the user decision to charge is described as follows: The current SoC is at 50.4\%, which is sufficient for short-term operations but considering the time of day and the need for operational readiness, it is prudent to charge now.

To ensure the scientific rigor and systematic approach of the analysis, this study adopted a method that integrates keyword extraction with sentence-by-sentence analysis. The method commences with the identification of the principal elements affecting user decision-making using keyword extraction.  The sentence-by-sentence analysis captures semantics and examines their interactions with the five components (SoC, Time, Range, Price, Psychology), thereby elucidating the specific influence of each element on user decision-making. Specifically, the first sentence indicated that although the SoC is sufficient for short-term operations, the user opted to charge in view of long-term operational requirements; this shows that the SoC is one of the factors considered in the decision-making process and is thus assigned a value of 1. The second sentence highlighted that time-windows and operational readiness directly prompted the charging decision, so they are also assigned a value of 1. Price and psychological factors are not mentioned in the text, and are therefore assigned a value of 0. This example demonstrates the interaction of many aspects in the charging decision-making process of EV users, affecting user behavior and offering crucial insights for optimizing charging techniques and improving user experience.

\bibliographystyle{ieeetr.bst}
\bibliography{reference}

\end{document}